\documentclass[sigconf]{acmart} 
\title{REMOT: A Region-to-Whole Framework for Realistic Human Motion Transfer}

\usepackage{balance}
\usepackage{graphicx}
\usepackage{amsmath}
 
\usepackage{amssymb}
\usepackage{multirow}
\usepackage{makecell,rotating}
\usepackage{array}
\usepackage{booktabs}
\usepackage{lipsum}
\usepackage{setspace}

\newcommand\blfootnote[1]{% 
\begingroup 
\renewcommand\thefootnote{}\footnote{#1}% 
\addtocounter{footnote}{-1}% 
\endgroup 
}

\def\BibTeX{{\rm B\kern-.05em{\sc i\kern-.025em b}\kern-.08emT\kern-.1667em\lower.7ex\hbox{E}\kern-.125emX}}

\copyrightyear{2022} 
\acmYear{2022} 
\setcopyright{acmcopyright}\acmConference[MM '22]{Proceedings of the 30th ACM International Conference on Multimedia}{October 10--14, 2022}{Lisboa, Portugal}
\acmBooktitle{Proceedings of the 30th ACM International Conference on Multimedia (MM '22), October 10--14, 2022, Lisboa, Portugal}
\acmPrice{15.00}
\acmDOI{10.1145/3503161.3547896}
\acmISBN{978-1-4503-9203-7/22/10}

\begin{document}

%\fancyhead{}

% \title{Progressive Region-to-Whole GAN for Human Motion Transfer}
\title{REMOT: A Region-to-Whole Framework for Realistic Human Motion Transfer}


% \author{Quanwei Yang$^{1,2}$, Xinchen Liu$^1$, Wu Liu$^{1*}$, Hongtao Xie$^{2*}$, Xiaoyan Gu$^3$,\\ Lingyun Yu$^{2,4}$, Yongdong Zhang$^2$}
% \affiliation{
%   \institution{$^1$JD Explore Academy,
%   $^2$University of Science and Technology of China,
%   $^3$Institute of Information Engineering, Chinese Academy of Sciences, 
%   $^4$Institute of Artificial Intelligence, Hefei Comprehensive National Science Center}
% }
% \email{{yangquanwei,yuly}@mail.ustc.edu.cn,{liuxinchen1,liuwu1}@jd.com, {htxie,zhyd73}@ustc.edu.cn, guxiaoyan@iie.ac.cn}

\author{Quanwei Yang}
\email{yangquanwei@mail.ustc.edu.cn}
\authornotemark[2]
\affiliation{%
  \institution{University of Science and Technology of China}
}

\author{Xinchen Liu}
\authornotemark[1]
\email{liuxinchen1@jd.com}
\author{Wu Liu}
\email{liuwu1@jd.com}
\affiliation{%
  \institution{JD Explore Academy}
}

% \author{Wu Liu}
% \email{liuwu1@jd.com}
% %\orcid{1234-5678-9012}
% \affiliation{%
%   \institution{JD Explore Academy}
% %   \streetaddress{P.O. Box 1212}
% %   \city{Dublin}
% %   \state{Ohio}
% %   \country{USA}
% %   \postcode{43017-6221}
% }

\author{Hongtao Xie}
\email{htxie@ustc.edu.cn}
\affiliation{%
  \institution{University of Science and Technology of China}
}

\author{Xiaoyan Gu}
\email{guxiaoyan@iie.ac.cn}
\affiliation{%
  \institution{Institute of Information Engineering, Chinese Academy of Sciences}
}

\author{Lingyun Yu}
\email{yuly@mail.ustc.edu.cn}
\affiliation{%
  \institution{University of Science and Technology of China}
  \institution{Institute of Artificial Intelligence, Hefei Comprehensive National Science Center}
}

\author{Yongdong Zhang}
\email{zhyd73@ustc.edu.cn}
\affiliation{%
  \institution{University of Science and Technology of China}
}

\renewcommand{\shortauthors}{Quanwei Yang et al.}

\begin{abstract}
Human Video Motion Transfer (HVMT) aims to, given an image of a source person, generate his/her video that imitates the motion of the driving person. 
Existing methods for HVMT mainly exploit Generative Adversarial Networks (GANs) to perform the warping operation based on the flow estimated from the source person image and each driving video frame. 
However, these methods always generate obvious artifacts due to the dramatic differences in poses, scales, and shifts between the source person and the driving person. 
To overcome these challenges, this paper presents a novel \textbf{RE}gion-to-whole human \textbf{MO}tion \textbf{T}ransfer (\textbf{REMOT}) framework based on GANs. 
To generate realistic motions, the REMOT adopts a progressive generation paradigm: it first generates each body part in the driving pose without flow-based warping, then composites all parts into a complete person of the driving motion.
Moreover, to preserve the natural global appearance, we design a Global Alignment Module to align the scale and position of the source person with those of the driving person based on their layouts. 
Furthermore, we propose a Texture Alignment Module to keep each part of the person aligned according to the similarity of the texture.
Finally, through extensive quantitative and qualitative experiments, our REMOT achieves state-of-the-art results on two public benchmarks.

\end{abstract}




%%\settopmatter{printfolios=true}
\maketitle

% \begin{small}
% \begin{spacing}{1}
% \textbf{ACM Reference Format:}
% \blfootnote{${\dagger}$This work is done when Quanwei Yang is an intern at JD Explore Academy.}
% \blfootnote{*Wu Liu and Hongtao Xie are the corresponding authors.}

% \noindent Quanwei Yang, Xinchen Liu, Wu Liu, Hongtao Xie, Xiaoyan Gu, Lingyun Yu, Yongdong Zhang. 2022. REMOT: A Region-to-Whole Framework for Realistic Human Motion Transfer. In \textit{Proceedings of the 30th ACM International Conference on Multimedia (MM'22), October 21–25, 2022, Lisbon, Portugal}. ACM, New York, NY, USA, 
% \pageref{LastPage} pages. \url{https://doi.org/10.1145/3343031.3350857}
% \end{spacing}
% \end{small}

\begin{small}
\begin{spacing}{1}
\blfootnote{${\dagger}$This work is done when Quanwei Yang is an intern at JD Explore Academy.}
\blfootnote{*Xinchen Liu is the corresponding author.}
\end{spacing}
\end{small}

\begin{figure}[t]
  \centering
  \includegraphics[width=\linewidth]{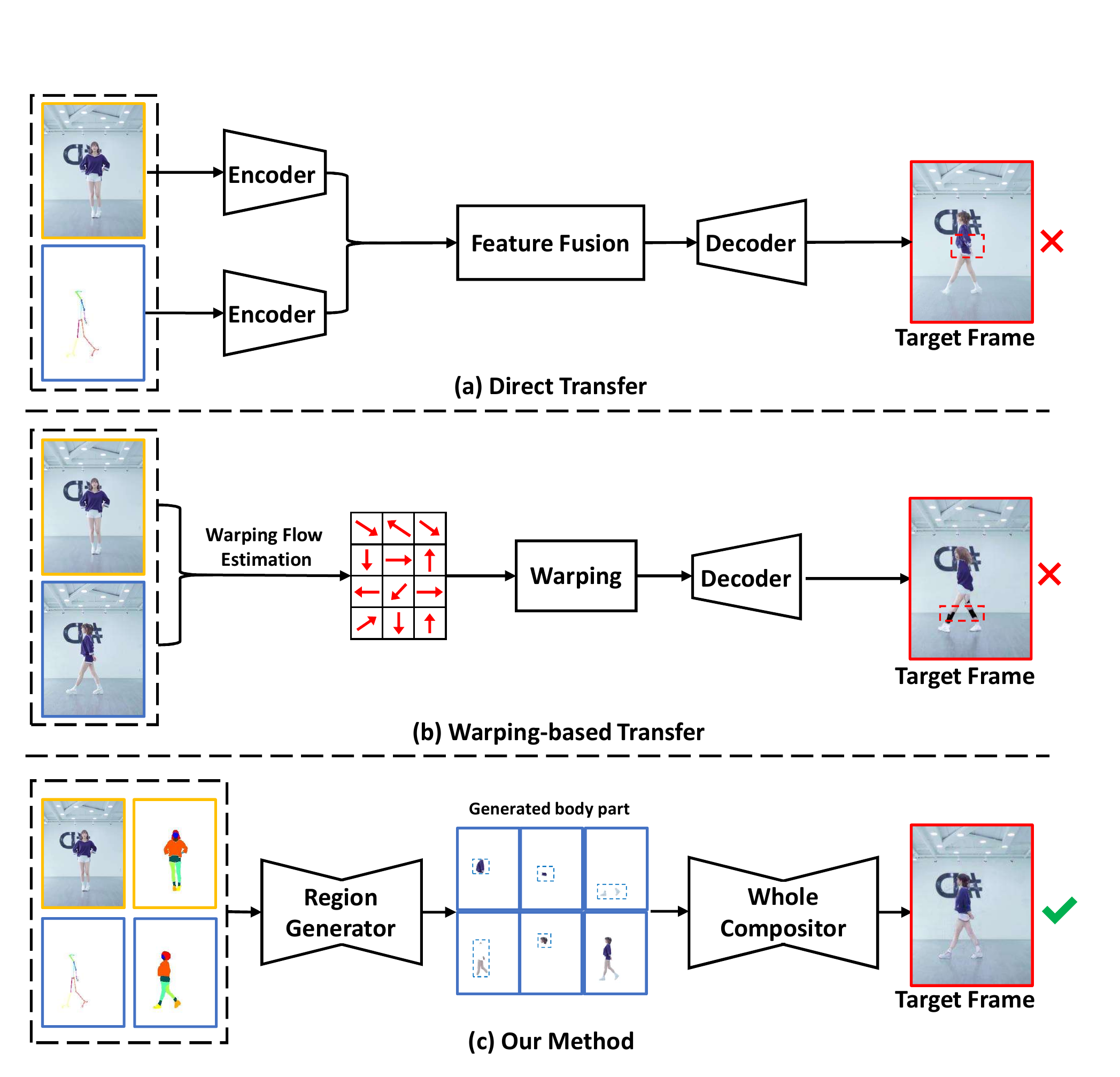}
  \caption{Comparison of our method with direct transfer methods and warping-based transfer methods. The images with the \textcolor{orange}{orange} box are the source image and human parsing. The images with the \textcolor{blue}{blue} box represent the driving video frame, pose map and human parsing. The images with the \textcolor{red}{red} box are the generated frames. Obvious artifacts in the generated images are marked with red dotted lines.}
  \label{fig_intro} 
  \vspace{-5mm}
\end{figure}

\section{Introduction}
The task of Human video Motion Transfer (HVMT) is to synthesize a realistic video where a source person performs desired driving motion~\cite{MaJSSTG17,c2f}. 
Specifically, given the source image and a driving video, it needs to make the person in the source image imitate the action of the person in the driving video. 
In recent years, this task has received intense research because of its wide range of applications, such as digital man~\cite{dig} and person re-identification~\cite{re-id,gait}.
However, the quality of generated motion videos is still far from satisfactory and needs further improvement.
% Therefore, it is of great importance to  the research on Human video Motion Transfer.

With the development of deep learning, remarkable achievements have been made in many fields~\cite{GoodfellowPMXWOCB14,transformer,pymaf,social,putting,mask-rcnn,yolo,fcn,human_rec,img2word2,pose_pre,resnet,action}. Among them, the rise of Generative Adversarial Networks (GANs)~\cite{GoodfellowPMXWOCB14} makes synthesized images or videos become more and more realistic~\cite{Wang0ZTKC18,v2v,face1,face2,stylegan}.
% Due to the rapid development of Generative Adversarial Networks (GANs)~\cite{GoodfellowPMXWOCB14}, synthesized images or videos are becoming more and more realistic~\cite{Wang0ZTKC18,v2v,face1,face2}.
Existing HVMT methods also adopt the GAN-based framework to achieve realistic results~\cite{edn}.
%Therefore, some works in HVMT have achieved relatively successful results. 
According to the generalization of the model, current HVMT methods can be divided into personalized HVMT and general-purpose HVMT. 
Personalized methods~\cite{edn, WangCSM21, YangZW00ZL20} implicitly embed the global appearance of a specific person into the generator, then synthesize the images of this person conditioned on the driving poses.
Although such methods can synthesize realistic video frames, these methods require a large amount of video data of a specific person for training. 
%When we want to synthesize another person's videos, we still need to collect video data of this person and train the model again.
Moreover, this type of method has no generalization ability for different persons, which has to train an exclusive model for each subject.

To overcome the huge data requirement of individual subjects and improve the generalization of personalized methods, recent works mainly focus on general-purpose methods~\cite{lwgan,c2f,SiarohinLT0S19,JeonNOK20,Wang0TLCK19,HuangHXZ21,GafniAW21}.
%Once the model of a general approach is trained, it can also be used for unseen person generation. Even if generated images on unseen people are unsatisfactory, we only need to fine-tune the existing model instead of retraining the model. 
This type of method aims to learn a model that can be adapted for the generation of unseen persons.
Even if the generated images may be not satisfactory, the results can be improved by simply fine-tuning the existing models instead of re-training them.
Therefore, this paper is also concentrated on the general-purpose human motion transfer which has a wider range of applications.

Although general-purpose methods have many advantages, they still face several challenges. 
First of all, because different parts of one person, such as the face, the body, the pants, etc., usually have different textures, it is difficult to model these complex textures with a uniform GAN.
For example, some direct transfer methods~\cite{Wang0TLCK19} use a GAN to generate the whole person by fusing the appearance and pose features, which hardly preserve local details well, as shown in Figure~\ref{fig_intro} (a).
Second, the drastic variations of poses and viewpoints between the source person and the driving person bring great challenges to human motion transfer.
Some recent methods adopt warping flows, such as the optical flow~\cite{c2f}, the transform flow~\cite{lwgan}, etc., to achieve pose transfer, as shown in Figure~\ref{fig_intro} (b).
Nevertheless, the generated frames still have significant artifacts caused by the errors of flow estimation.
Last but not the least, the misalignment of scales and positions between the source person and the driving person can also degrade the quality of generated videos.

To overcome the above challenges, this paper presents a novel \textbf{RE}gion-to-whole framework for realistic human \textbf{MO}tion \textbf{T}ransfer, named REMOT.
Figure~\ref{fig_intro} (c) demonstrates the conceptual architecture of the proposed progressive framework.
The REMOT first predicts the layout (semantic parsing masks) of the human body conditioned on the driving pose.
Then a region generator is designed to generate each body part individually.
Moreover, we propose a Global Alignment Module (GAM) to keep the scales and positions of the source person and the driving person aligned for accurate feature fusion.
Finally, a GAN-based whole compositor is utilized to integrate the parts and refine the coarse synthesized person image.
To preserve the local details of the source person, we design a Texture Alignment Module (TAM) to align the features of the source person image and the generated frames.
% It should be emphasized that our method abandons the complex warping flow estimation and only uses GANs to generate person images, so our model can better deal with dramatic difference of poses. Because our model gradually generates person images from simple to complex, our model training is more stable.
In summary, our key contributions are as follows:
\begin{itemize}
	\item We propose a region-to-whole framework, named REMOT, for the general-purpose HVMT task. REMOT discards the flow-based warping operation and takes a progressive generation paradigm, which can better deal with dramatic variations of poses by gradually generating person frames from parts to the whole.
	\item To overcome the differences in scales and positions between the source person and the driving ones, we propose a Global Alignment Module to adapt the scale and position of the source person to those of the driving person.
	\item A Texture Alignment Module is proposed to align the features of the source image and the initially generated image to preserve more details like textures of clothes and edges of the bodies.
% 	\item[$\bullet$] We conduct extensive experiments on the iPER Dataset\cite{lwgan} and SoloDance Dataset\cite{c2f}. Experimental results show that our model achieves state-of-the-art both quantitatively and qualitatively.
\end{itemize}
We conduct extensive experiments on the iPER Dataset~\cite{lwgan} and SoloDance Dataset~\cite{c2f}. 
Experimental results show that our model achieves the state-of-the-art  both quantitatively and qualitatively.

% \begin{equation}
% \begin{aligned}
% X_{S} &= \begin{cases}{\left[x_{(t+1)}, x_{(t+2)}, \ldots, x_{(t+S)}\right],} & L>=S, \\
% \operatorname{Sort}\left(\left[x_{1}, x_{2}, \ldots, x_{L}, X_{1}, \ldots, X_{(S-L)}\right]\right), & L<S,\end{cases} \\
% &= \begin{cases}{\left[x_{(t+1)}, x_{(t+2)}, \ldots, x_{(t+S)}\right],} & L>=S, \\
% {\left[x_{1}, X_{1}, \ldots, X_{(S-L)}, X_{(S-L)}, X_{(S-L+1)}, \ldots, x_{L}\right],} & L<S,\end{cases}
% \end{aligned}
% \end{equation}

\section{Related Work}
%相关工作
\subsection{Image-to-Image Translation}
Our method is related to the image-to-image translation task that converts one source image to a target image conditional on a specific style, resolution, contents, etc~\cite{IsolaZZE17}.
Existing image-to-image translation methods usually adopt the Conditional Generative Adversarial Networks~\cite{MirzaO14,LiCWZ19} to take the specific image (e.g., semantic segmentation map, sketch, edge map, etc.) as input for image generation. 
Such models can generate the desired image by editing the properties of the input image. 
Therefore, Compared with unconstrained GANs, image-to-image translation has a wider range of applications.
For example, Pix2pix~\cite{IsolaZZE17} used a conditional discriminator to make the generated image and the input image match as much as possible, so it can generate realistic corresponding images. 
Based on pix2pix, pix2pixHD~\cite{Wang0ZTKC18} generated high-resolution images by using multi-scale generators and discriminators. 
Vid2vid~\cite{v2v} was aimed at the temporal inconsistency of the generated videos. 
By adding optical flow constrained to the generator and discriminator, vid2vid can generate temporally coherent videos. 

The personalized motion transfer is to translate the pose map to the corresponding RGB image, where the person's appearance is implicitly embedded into the generator.
Therefore, many image-to-image translation works can be used for personalized motion transfer.
% But ignoring the characteristics of HVMT, such methods sometimes generates unreasonable person images, such as blurred faces.
However, such methods cannot be directly applied to HVMT for the generation of varied poses and precise details of persons.

\begin{figure*}[t]
  \centering
  \includegraphics[width=0.98\linewidth]{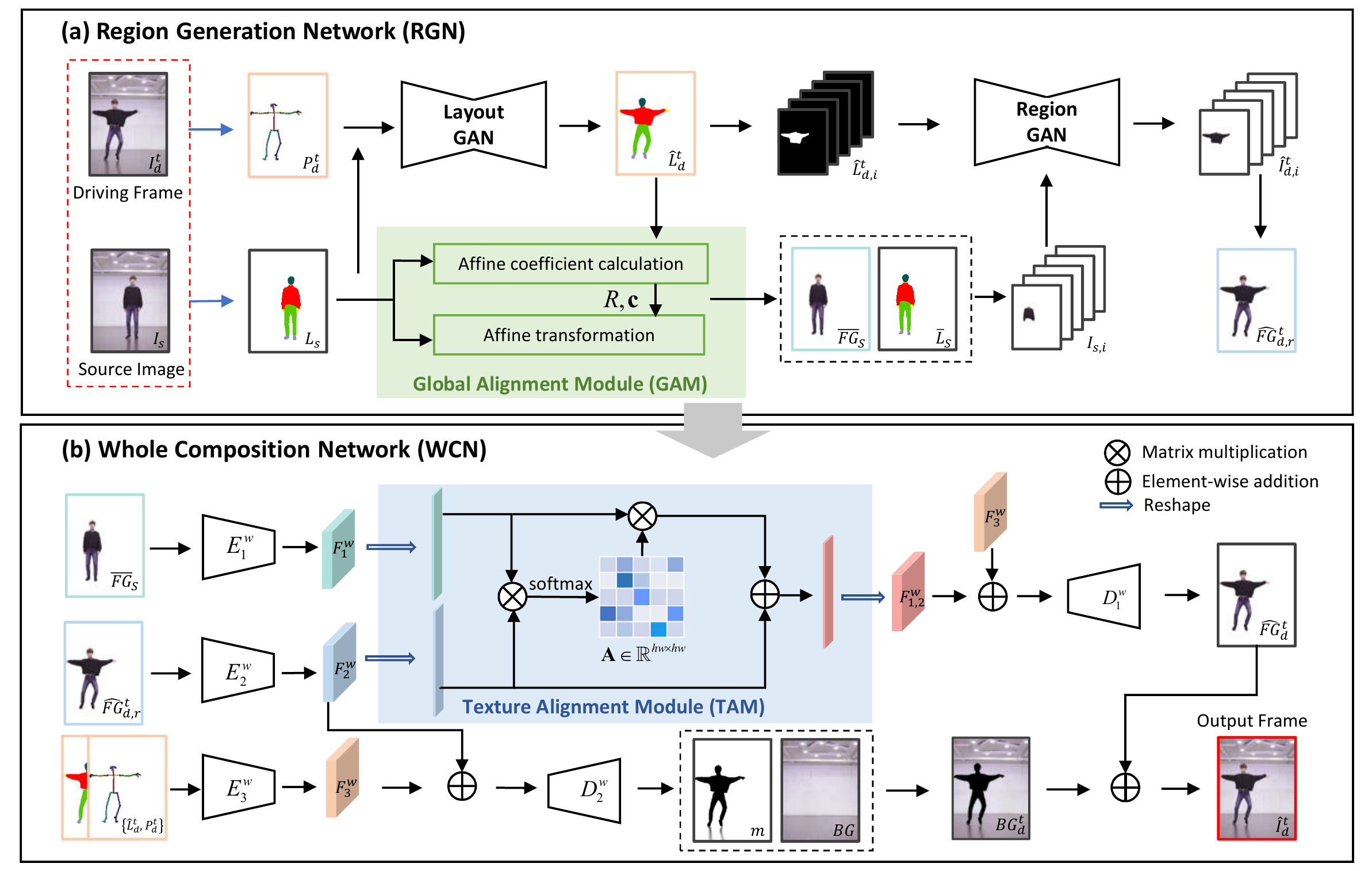}
%   \caption{The pipeline of our proposed method. The images inside red dotted box are the input of the model, and the blue arrows represent the data preprocessing: pose detection, layout detection. Our method consists of two networks. First, The Region Generation Network (RGN) generates the person region images in the driving pose according to the generated layout and the aligned region images of source people with the Global Alignment moudle (GAM). Then the Whole Composition Network (WCN) integrates the generated region images, further refines the texture details with the Texture Alignment module (TAM), and adds a background to the generated person image. (The images with the red box are the input of the WCN. Best viewed in color.)}
   \caption{The pipeline of our proposed REMOT. It consists of two networks: (a) Region Generation Network (RGN) and (b) Whole Composition Network (WCN). The RGN takes the source image and driving frame as input to obtain each  region of the source person  in the driving pose. (The blue arrows represent the data preprocessing: pose detection, layout detection.) The WCN takes the output of the RGN $\hat{FG}_{d,r}^t$, the source foreground image after global alignment $\overline{FG}_{s}$, and concatenated human layout with pose map $\{\hat{L}_d^{t}, P_d^{t}\}$ as input, then generates the final target frame.}
 \label{pipe} %\vspace{-2mm}
\end{figure*}

\subsection{Human Video Motion Transfer}
\textbf{Personalized HVMT.} 
On the basis of the image-to-image translation methods, some works have specific improvements based on the characteristics of this task. 
In order to conform to the size of the source person, EDN~\cite{edn} matched the poses between the driving person and the source person by designing global pose normalization. 
In addition, EDN designed a face GAN to generate a more precise face region. 
DIW~\cite{WangCSM21} encoded non-uniformly sampled pose maps of the past ten frames to capture appearance change details and refine pose features of the single frame. 
As mentioned earlier, personalized methods suffer from the inability to generalize to unseen person generations and need massive computing resources.

\textbf{General-purpose HVMT.} 
To overcome the limitations of personalized HVMT, recent studies focus on general-purpose HVMT.
In order to better preserve appearance details, some works adopt the warp operation.
% For example, FOMM~\cite{SiarohinLT0S19}and MRAA~\citep{mr} computed the flow field  between the source image and the driving image, and then warped the feature of the source image.
These methods focusing on the animation of general objects, such as FOMM~\cite{SiarohinLT0S19} and MRAA~\citep{mr}, utilized predicted keypoints or regions to compute the flow field between the source image and the driving image, and then warped the feature of the source image to obtain the target image.
However, these methods do not take temporal consistency into consideration, which results in the temporal inconsistency of the synthesized video. 
LWGAN~\cite{lwgan} warped the source information in both image and feature spaces based on the SMPL model. 
However, the SMPL model~\cite{smpl} was only suitable for smooth human bodies, it cannot represent the human body with complex clothes. 
Different from warping the feature of the source image, C2F~\cite{c2f} estimated the optical flow of the clothing regions and directly warped the clothing region according to optical flow. 
Unfortunately, when the driving pose is greatly different from the source pose, such methods usually failed due to inaccurate flow estimations. 
Different from the above methods, our proposed method does not use warp operation. 
This design can avoid complex transformation calculations and can better cope with arbitrary poses. 

In addition, some methods consider the diversity of different regions of the human body and treat each region of the human body separately.
SSF~\cite{GafniAW21} encoded each region of the body separately, then integrated features of different regions to directly generate the final target image. 
SHUP~\cite{SIHUP} first transformed each part of the source person to the spatial position of the driving pose, then directly generated the whole person image.
Moreover, PGTM~\cite{HuangHXZ21} predicted the UV map~\cite{uv} of the source person conditioned on driving pose, then filled it with each region textures of the source person image. 
However, predicting dense UVs in different poses accurately is challenging. 
StylePeople~\cite{StylePeople} used parameterized human body model SMPL-X~\cite{smpl-x} to model the pose and shape of subjects, then used neural rendering to render region appearance details.
Our method also encodes each region of the source image separately. 
However, unlike these methods, we directly generate corresponding target region images, then use the whole composition network to integrate the generated regions for generating the final target image. 
Compared with directly generating the whole person image, generating the region image first can better preserve the appearance details.

\section{The Proposed Method}
Given the source image $I_s$ and a driving video $V_D=\{I_d^t\},t=1,2, \cdots, N$, where $I_d^t$ represents the driving frame at time $t$ of the driving video, our goal is to generate a realistic video in which the person of the source image imitates the same motion as the driving video. We can formulate this task as

\begin{equation}
\hat{I}^t_{d,s} = \mathcal{F}\left(I_d^t, I_s \right), t=1,2, \cdots, N,
\end{equation}
where $\mathcal{F}(\cdot ,\cdot)$ refers to generative model, $N$ represents the number of video frames, $\hat{I}^t_{d,s}$ represents the generated image.

Figure~\ref{pipe} is an overview of our pipeline. Our proposed method consists of two networks: the Region Generation Network and the Whole Composition Network.
For the Region Generation Network, given the pose of the driving image $P_d^t$ and the human body layout of the source image $L_s$, we first generate the layout of the source person conditioned on the driving pose $\hat{L}_d^t$ through Layout GAN. Then the Region GAN takes the generated mask of each region $\hat{L}_{d, i}^t$ and the corresponding region of the source person after global alignment $I_{s, i}$ as input, and obtains each source person region image $\hat{I}_{d, i}^t$ in the driving pose. Finally, in the Whole Composition Network, we refine the coarse person image $\hat{FG}_{d,r}^t$ by aligning the features of the source person with the Texture Alignment Module. In addition, we add the background image to the generated person image to obtain the final target image $\hat{I}_d^t$. In the following, we will introduce the details of these two networks.

% First, the layout generation module inputs the pose of the driving image $P_d^t$ and the human body layout of the source image $Ls$, and outputs the layout of the source person conditioned on the driving pose $\hat{L}_d^t$. The foreground image of the source image is then globally aligned according to the generated human body layout and the human body layout of the source image. In the region generation module, the generated mask of each region of the human body $\hat{L}_{d,i}^t$ and the RGB image of the corresponding region of the source person image after global alignment $I_{s,i}$ are as inputs to obtain and the RGB image of each region with a specific outline $\hat{I}_{d,i}^t$. By adding generated region images one by one, we can get the coarse foreground image of the source image conditioned on the driving pose $\hat{FG}_{d,r}^t$. Finally, in Stage3, by further integrating the information of the source image using the Texture Similarity-based Feature Fusion module, we refine the foreground image generated by stage2. In addition, we add the background image to the generated foreground image to obtain the final generated image $\hat{I}_d^t$. In the following, we will introduce the details of each stage.

\subsection{Region Generation Network}
As mentioned before, the reason why we first generate each source person region in the driving pose is that different regions of the human body have significantly different texture patterns. It's difficult to directly generate the whole human body. To achieve this, we need to first generate the layout of the source people in the driving pose through Layout GAN, as shown in Figure~\ref{pipe} (a).
% As in previous work\cite{c2f}, the reason why we first generate the layout of the source person conditioned on the driving pose is that it is easier to generate the layout than to directly generate the corresponding RGB image.

\textbf{Layout GAN.} In order to represent the human pose,  we first use the human keypoint detection method~\cite{openpose,pose2,pose3,liu2021} to predict the 2-D keypoint $K\in\mathbb{R}^{2 \times 25}$ of the human body. Then according to the predefined connection strategy, we can get the pose connection map $P_d^t\in\mathbb{R}^{3\times H\times W} $ of the driving person, where $H \times W$ is the resolution of the image. 
In order to represent the human body layout, we use the human parsing methods~\cite{layout,braid,NAS} to obtain the 18-channel semantic segmentation maps of the person image. Further, according to the similarity of textures in various regions, we merge the semantic segmentation maps into six classes, i.e., head, top, bottom, shoes, limbs, and background. So we can get the body layout $L_s\in \mathbb{R}^{6\times H \times W} $ of the source person image and $L_d^t$ of the driving person image.

For the network structure, we also use vid2vid~\cite{v2v} as in \cite{c2f}. %Vid2vid is a classic framework for video-to-video translation, which is widely used because of its excellent generation results. 
Specifically, we use three encoders $E_1^l$, $E_2^l$ and $E_3^l$ to encode the source body layout $L_s$, the concatenated driving poses connection map $\{P_d^{t-2}, P_d^{t-1}, P_d^t\}$ and the concatenated results previously generated $\{\hat{L}_d^{t-2},\hat{L}_d^{t-1}\}$, to obtain the feature maps $F_1^l$, $F_2^l$ and $F_3^l$ respectively. Then decoder $D_1^l$ is proposed to decode the added feature maps $F_1^l$, $F_2^l$ and $F_3^l$ to get raw result $\hat{L}_{d,raw}^{t}$. Similarly, $D_2^l$ decodes added features $F_2^l$ and $F_3^l$ to obtain optical flow $O$ and its weight $w$, and finally obtains the final result $\hat{L}_d^{t}$. The entire Layout GAN can be formulated as
\begin{subequations}
\begin{align}
&\hat{L}_{d,raw}^{t} = D_1^l\left(E_1^{l}(L_s)+E_2^{l}(\{P_d^{t-2}, P_d^{t-1}, P_d^t\})+E_3^{l}(\{\hat{L}_d^{t-2},\hat{L}_d^{t-1}\})\right),\\
&O,w = D_2^l\left(E_2^{l}(\{P_d^{t-2}, P_d^{t-1}, P_d^t\})+E_3^{l}(\{\hat{L}_d^{t-2},\hat{L}_d^{t-1}\})\right),\\
&\hat{L}_d^{t} = \hat{L}_{d,raw}^{t}*w + \operatorname{Warp}(\hat{L}_d^{t-1},O)*(1-w),
\end{align}
\end{subequations}
where $+,*$ represent element-wise addition and element-wise multiplication, respectively. $\{\}$ indicates that the inputs are concatenated along the channel dimension. $\operatorname{Warp}(I , o)$ refers to the affine transformation of $I$ according to the optical flow $O$. It should be emphasized that the warp operation in the formula is only to integrate the generated results from previous moment and make generated video frames coherent temporally, not for the driving image and the source image.

\textbf{Region GAN.} After obtaining the human body layout $\hat{L}_d^t$ of the source person conditioned on driving pose from Layout GAN, Region GAN is proposed to generate five person regions conditioned on the driving pose.

For the network structure, we still adopt the vid2vid structure. Since Region GAN is only for generating raw region images of the source person, here we only use one generator to generate five regions of the human body. This can not only save computational resources but also prevent the model from over-fitting. Specifically, we use the encoder $E_1^r$ to encode the $i$-th region mask $\hat{L}_{d, i}^t\in\mathbb{R}^{1\times H\times W}$ of the human body to obtain the feature map $F_1^r$. In addition, we use another encoder $E_2^r$ to encode the $i$-th region of the source person image and the previous generations to obtain the feature map $F_2^r$. Here we directly concatenate $\hat{I}_{d, i}^{t-1}$, $\hat{I}_{d, i}^{t-2}$and $I_{s, i}$ along the channel dimension to utilize the previous generations for enhancing the temporal consistency of appearance details. The original Region GAN can be expressed as
\begin{equation}
\hat{I}_{d,i}^{t} = D_1^r\left(E_1^{r}(\hat{L}_{d,i}^t)+E_2^{r}(\{\hat{I}_{d,i}^{t-1},\hat{I}_{d,i}^{t-2}, I_{s,i}\})\right).
\end{equation}

\textbf{Global Alignment Module.} The scale and spatial position of the person in the video may vary greatly, so the scales and positions of $\hat{I}_{d, i}^{t}$ and $I_{s,i}$ may not match, then the extracted features $F_1^l$ and $F_2^l$ are also not aligned. Therefore, simply adding $F_1^l$ and $F_2^l$ does not fuse the features well. To scale it, we propose the Global Alignment Module (GAM), a simple but effective Affine transformation to transform the $I_s$ before feeding it to $E_2^r$, so that the scale and location of $I_s$ are consistent with the $L_d^t$.

%%%Global alignment

See the green rectangle of Figure~\ref{pipe} (a). First, we obtain the human masks $M_d^t$ and $M_s$ according to $\hat{L}_d^t$ and $L_s$, respectively. According to the human mask, we can obtain the foreground image ${FG}_s$ of the source image. We express the process of aligning the foreground of the source foreground image with the following formulation:
\begin{equation}
\overline{FG}_{s} = \left[\begin{array}{ll}
R & 0 \\
0 & R
\end{array}\right] FG_{s}+\left[\begin{array}{c}
c_{x} \\
c_{y}
\end{array}\right],
\end{equation}
where $\overline{FG}_{s}$ represents the global aligned foreground image of the source image. $R$ represents the scaling factor, which is obtained according to $H_d/H_s$, $H_d$ and $H_s$ represent the height of $M_d$ and $M_s$, respectively. ${\bf{c}}=[c_x,c_y]^T$ represents the position offset, which can be obtained by $\bf{c}_d-\bf{c}_s$, where $\bf{c}_d$ and $\bf{c}_s$ represent the center position coordinates of $M_d^t$ and $M_s$ , respectively. So the Region GAN can be expressed as
\begin{equation}
\hat{I}_{d,i}^{t} = D_1^r\left(E_1^{r}(\hat{L}_{d,i}^t)+E_2^{r}(\{\hat{I}_{d,i}^{t-1},\hat{I}_{d,i}^{t-2}, \operatorname{GAM}(I_{s,i})\})\right).
\end{equation}

The loss of Region GAN is the sum of the loss of five synthesized region images, and each loss consists of three items, e.g., the reconstruction loss, the perceptual loss~\cite{lossp}, and the adversarial loss.

\textbf{Reconstruction loss.} It directly constrains the generated image $\hat{I}_{d,i}^{t}$ in the pixel space to be closer to ground truth ${I}_{d,i}^{t}$. Compared with the L2 loss, L1 loss can focus on the subtle differences of the image. Its formulation is as follows:
\begin{equation}
\mathcal{L}_{rec,i}^r=\left\|\hat{I}_{d,i}^{t}-{I}_{d,i}^{t}\right\|_{1}.
\end{equation}

\textbf{Perceptual loss.} It regularizes the generated image $\hat{I}_{d, i}^{t}$ and ground truth ${I}_{d, i}^{t}$ to be closer in multi-dimensional feature space. The perceptual loss includes the feature content loss and feature style loss, which can be expressed as
\begin{equation}
\mathcal{L}_{per,i}^r=\sum\limits_{j=1}^{n}\left(\left\|\phi_j\left(\hat{I}_{d,i}^{t}\right)-\phi_j\left({I}_{d,i}^{t}\right)\right\|_{1} + \left\|G_j^\phi\left(\hat{I}_{d,i}^{t}\right)-G_j^\phi\left({I}_{d,i}^{t}\right)\right\|_{1}\right),
\end{equation}
where $j$ represents the $j$-th layer of the pre-trained VGG-19~\cite{vgg} model, and $G$ represents the Gram matrix of features.

\textbf{Adversarial loss.} It aims to force the synthesized images to conform to a similar distribution of real images. In order to make the network focus on multi-scale image details, we also use the multi-scale conditional discriminator proposed in pix2pixHD~\cite{IsolaZZE17}. It takes a synthesized image and the corresponding human region mask as input. Its expression is
\begin{equation}
\mathcal{L}_{Gadv,i}^r=\sum D_k(\hat{I}_{d,i}^{t},L_{d,i}^t).
\end{equation}
Therefore, the entire loss function of the generator is shown below.
\begin{equation}
\mathcal{L}_{G}^r =\sum\limits_{i=1}^{5}\left(\lambda_{rec}\mathcal{L}_{rec,i}^r+\lambda_{per}\mathcal{L}_{per,i}^r+\mathcal{L}_{Gadv,i}^r\right),
\end{equation}
where $\lambda_{rec}$ and $\lambda_{per}$ are the weights for the reconstruction loss and perceptual loss, respectively.

For the discriminator, its loss function is
\begin{equation}
\mathcal{L}_{D}^r=\sum \mathbb{E}[\log D_k({I}_{d,i}^{t},L_{d,i}^t)]+\mathbb{E}[\log (1-D_k(\hat{I}_{d,i}^{t},L_{d,i}^t))].
\end{equation}

\subsection{Whole Composition Network}
After generating the each region image, we can obtain the foreground image of the target image using the following formula:
\begin{equation}
\hat{FG}_{d,r}^t=\sum\limits_{i=1}^{5} \hat{I}_{d,i}^t .
\end{equation}
However, this is only the sum of the individual regions of the human body generated by Region GAN, each region is still insufficient in connections and some details. 
In addition, we still need to add the background to the generated person image. 
Therefore, the Whole Generation Network is proposed to further enhance the details of the person, improve the unreasonable areas and add the background image to the person image.

As shown in Figure~\ref{pipe} (b), for the Whole Composition Network, we use three encoders to extract information. a) We use encoder $E_3^w$ to encode the concatenated $\{\hat{L}_d^{t-2\sim t},P_d^{t-2\sim t}\}$ to obtain feature map $F_3^w\in\mathbb{R}^{c\times h\times w}$.
Three adjacent frames as input can implicitly improving the temporal consistency of generated frames.
b) We use the encoder $E_2^w$ to encode concatenated $\{\hat{FG}_{d}^{t-2},\hat{FG}_{d}^{t-1},\hat{FG}_{d,r}^t\}$ to get the feature map $F_2^w\in\mathbb{R}^{c\times h\times w}$. 
c) In order to exploit the true foreground information of the source person image again, we use the encoder $E_1^w$ to encode $\overline{FG}_s$ to get the feature map $F_1^w\in\mathbb{R}^{c\times h\times w}$. Because $\hat{L}_d^{t}$, $P_d^{t}$ and $\hat{FG}_{d,r}^t$ are all in the driving pose, the obtained feature maps $F_3^w$ and $F_2^w$ can be directly added for feature fusion. 

\textbf{Texture Alignment Module.} However, $\hat{FG}_{d,r}^t$ and $\overline{FG}_{s}$ are the same person with the same textures but having different poses. So the Texture Alignment Module (TAM) is proposed to better fuse feature maps $F_1^w$ and $F_2^w$ for refining local details.
TAM calculates the texture similarity between each position of $F_1^w$ and each position of $F_2^w$, then adds the aligned $F_1^w$ to $F_2^w$. 

As shown in the blue rectangle of Figure~\ref{pipe} (b), we first reshape $F_1^w$ into $H_1\in\mathbb{R}^{c\times hw}$, and we can get $H_2\in\mathbb{R}^{c\times hw}$ by $F_2^w$ in the same way. Then we calculate the similarity of each position in $H_1$ to each position in $H_2$ to get the affinity matrix $\mathbf{A}\in\mathbb{R}^{hw\times hw}$. Here we use the cosine distance to represent the texture similarity. The calculation formula is as follows: 
\begin{equation}
\mathbf{A}=\text{softmax}\left(\frac{H_{1,i}^T\cdot H_{2,j}}{\|H_{1,i}\|\|H_{2,j}\|}\right),i,j=1,2,\dots,hw,
\end{equation}
where $\cdot$ means matrix multiplication, $H_{1,i}$ represents the feature of $H_1$ at the $i$-th position, and $H_{2,j}$ represents the feature of $H_2$ at the $j$-th position. Next, we multiply $H_1$ by the affinity matrix to get the alignment vector, then add it point-to-point with $H_2$, finally reshape it to the original dimension to get the fused feature map $F_{1,2}^w\in \mathbb{R}^{c\times h\times w}$, which can be formulated as
\begin{equation}
F_{1,2}^w=\text{reshape}\left(H_1\cdot \mathbf{A}+H_2\right).
\end{equation}
After getting the fused texture feature $F_{1,2}^w$, we add it to the pose feature $F_{3}^w$ and send it to decoder $D_1^w$ to get the final foreground image $\hat{FG}_{d}^t$. The entire foreground image generation process can be expressed as
\begin{equation}
\hat{FG}_{d}^t=D_1^w \left(F_{3}^w+\operatorname{TAM}(F_{1}^w,F_{2}^w)\right).
\end{equation}

Although a realistic person image is synthesized, we still need to add a reasonable background to the foreground person image. Therefore, we send the additive feature map $F_2^w$ and $F_3^w$ into decoder $D_2^w$ to get the soft-mask $m$, then multiply it with $BG$ to get the corresponding background image. Finally, we add the background to the foreground image to get the final image. Its calculation formula can be expressed as
\begin{equation}
\hat{I}_{d}^t=m*BG+\hat{FG}_{d}^t.
\end{equation}

It should be noted that we also use the same strategy as Layout GAN, using the predicted optical flow of adjacent frames to further refine the generated image. In addition to using the same loss function as Region GAN, we also design another discriminator specifically for the facial region to generate realistic faces.

\subsection{Training and Inference}
\textbf{Training.} In the training phase, in order to get the paired training data as supervision information, we select a forward frame in a video as the source image and this video as the driving video. 
%The reason for choosing the forward video frame as the source image is that it contains more appearance details of the person. 
Before training our model, we first use pre-trained models to extract the poses and the human parsing maps of frames. We train our REMOT step by step. First, we train Layout GAN and Region GAN separately for 10 epochs. Then we train the Whole Composition Network for 10 epochs using the output of Region GAN.

\textbf{Inference.} In the inference stage, the choice of driving video is not limited, as long as it is a clear video of any solo person. For convenience, our model can perform end-to-end inference.

\section{EXPERIMENTS}

This section consists of four subsections. We first introduce the Implementation Details and Datasets (Sec.~\ref{4.1}). Then compare our method with some representative existing methods quantitatively and qualitatively (Sec.~\ref{4.2} and Sec.~\ref{4.3}). Finally, we show the effectiveness of each module through the ablation study (Sec.~\ref{4.4}). Each subsection is described in detail below.

\begin{table}[t]
  \caption{Quantitative comparison to state-of-the-art methods on the iPER dataset.  $\uparrow$ indicates the higher is better, correspondingly,  $\downarrow$ indicates the lower is better.}
  \label{iper_1}
  \begin{tabular}{l||c|c|c|c|c}
    \hline
    Methods&SSIM $\uparrow$&PSNR $\uparrow$&LPIPS $\downarrow$&FID $\downarrow$&TCM $\uparrow$\\
    \hline
    \hline
    EDN~\cite{edn}   & 0.840 & 23.39  & 0.076 & 56.29  & 0.361 \\ \hline
    FSV2V~\cite{Wang0TLCK19} & 0.780 & 20.44  & 0.110 & 110.99 & 0.184 \\ \hline
    LWGAN~\cite{lwgan} & 0.825 & 21.43  & 0.091 & 77.99  & 0.197 \\ \hline
    C2F~\cite{c2f}   & 0.849 & 24.27  & 0.072 & 55.07  & 0.687 \\ \hline
    Ours    & \textbf{0.856}  &  \textbf{25.33} &  \textbf{0.065} &  \textbf{53.04} &  \textbf{0.793}  \\  \hline
\end{tabular}
\end{table}

\subsection{Implementation Details and Datasets}\label{4.1}
\textbf{Implementation Details.} Like~\cite{c2f}, all frames are resized and cropped to 256x192 before training the models. We use Openpose~\cite{openpose} to get the 2-D keypoints of the human body, use SCHP~\cite{layout} to get the human parsing, and use the optical flow obtained by FlowNet2~\cite{flow} as ground truth. We set $\lambda_{rec}$, $\lambda_{per}$ in Eq10 to 10 for balancing several losses. We use Adam optimizers~\cite{adam} with learning rate of 0.0002, where $\beta_1=0.5,\beta_2=0.999$ on Nvidia Tesla P40.

\textbf{Dataset.} In order to comprehensively evaluate the performance of our proposed model, we choose two public datasets for training and testing.

\textbf{iPER Dataset.}
The iPER dataset was proposed by~\cite{lwgan}, which was collected in the laboratory environment. The dataset includes 30 actors with a total of 206 videos. Its action types of an actor include A-pose and random actions. The scale of iPER is relatively large and the actions are relatively simple. Same as the original protocol of iPER, 164 videos are used for training and the remaining 42 videos for testing.

\textbf{SoloDance Dataset.} Different from the iPER dataset, SoloDance dataset contains 179 solo dance videos in real scenes collected online~\cite{c2f}, and each video has 300 frames. The dataset has 143 subjects, and its action types are mainly dances such as hip-hop and modern dance. Compared with the iPER dataset, the actions in this dataset are more complex and the types of clothes are more diverse. However, this dataset is not as large as the iPER dataset. Following the setting of~\cite{c2f}, we also use 153 videos for training and 26 videos for testing.

\textbf{Compared Approaches.}
We select several representative state-of-the-art HVMT methods for comparison with our proposed approach. They are personalized methods END~\cite{edn}, directly generation method FSV2V~\cite{Wang0TLCK19}, feature warping method LWGAN~\cite{lwgan}, and image warping method C2F~\cite{c2f}. We also use 3,000 video frames for each subject to train the exclusive EDN model, and the training strategies of other methods are the same as our method.

\begin{table}[t]
  \small
  \caption{Quantitative comparison to state-of-the-art methods on the SoloDance dataset. Here are Mask-SSIM and Mask-PSNR in ($\cdot$).}
  \label{solo_1}
  \begin{tabular}{l||c|c|c|c|c}
    \hline
    Methods&SSIM $\uparrow$&PSNR $\uparrow$&LPIPS $\downarrow$&FID $\downarrow$&TCM $\uparrow$\\
    \hline
    \hline
    EDN~\cite{edn}   & 0.811 & 23.22  & 0.051 & 53.17  & 0.347 \\ \hline
    FSV2V~\cite{Wang0TLCK19} & 0.721 & 20.84  & 0.132 & 112.99 & 0.106 \\ \hline
    LWGAN~\cite{lwgan} & 0.786 (0.935) & 20.87 (22.48)  & 0.106 & 86.53  & 0.176 \\ \hline
    C2F\cite{c2f}   & \textbf{0.879} (0.935) & \textbf{26.65} (24.94)  & 0.049 & \textbf{46.49}  & 0.641 \\ \hline
    Ours    & 0.850 (\textbf{0.953}) & 24.83 (\textbf{27.89}) & \textbf{0.045} & 53.29  & \textbf{0.788} \\  \hline
\end{tabular}
\end{table}

\subsection{Quantitative Results}\label{4.2}
\textbf{Metrics.} In order to comprehensively compare the quality of generated images, we use four image-level evaluation metrics: SSIM~\cite{ssim}, PSNR, Learned Perceptual Image Patch Similarity (LPIPS)~\cite{lpips}, Fréchet Inception Distance (FID)~\cite{fid} and one metric for evaluating the temporal consistency of generated videos, i.e., Temporal Consistency Metric (TCM)~\cite{tcm}. 
SSIM measures the structural similarity between the generated image and the real image in pixel space. PSNR evaluates the generation quality based on pixel-wise errors. LPIPS assesses the perceptual distance based on feature vectors. FID represents the Inception \cite{inception} feature distance between two image sets. As for TCM, it evaluates the temporal consistency of generated videos by calculating the distance between the current video frame and the warped neighbored frame with optical flow.

\begin{figure*}[t]
  \centering
  \includegraphics[width=\linewidth]{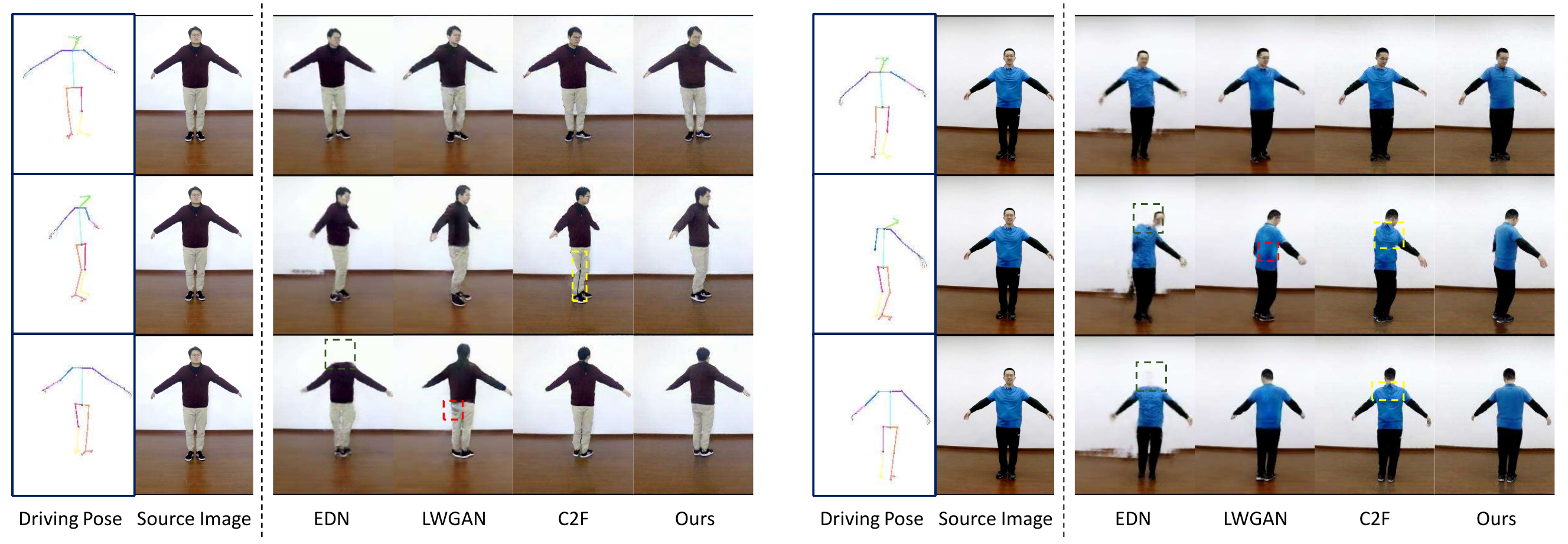}
  \caption{Qualitative results on the iPER dataset. 
  %To comprehensively compare the generated results, we select three generated images with different angles of the forward, side, and back in each video.
  Visible artifacts are marked with colored dotted boxes.}
  \label{iper} \vspace{-1mm}
\end{figure*}

\begin{figure*}[t]
  \centering
  \includegraphics[width=\linewidth]{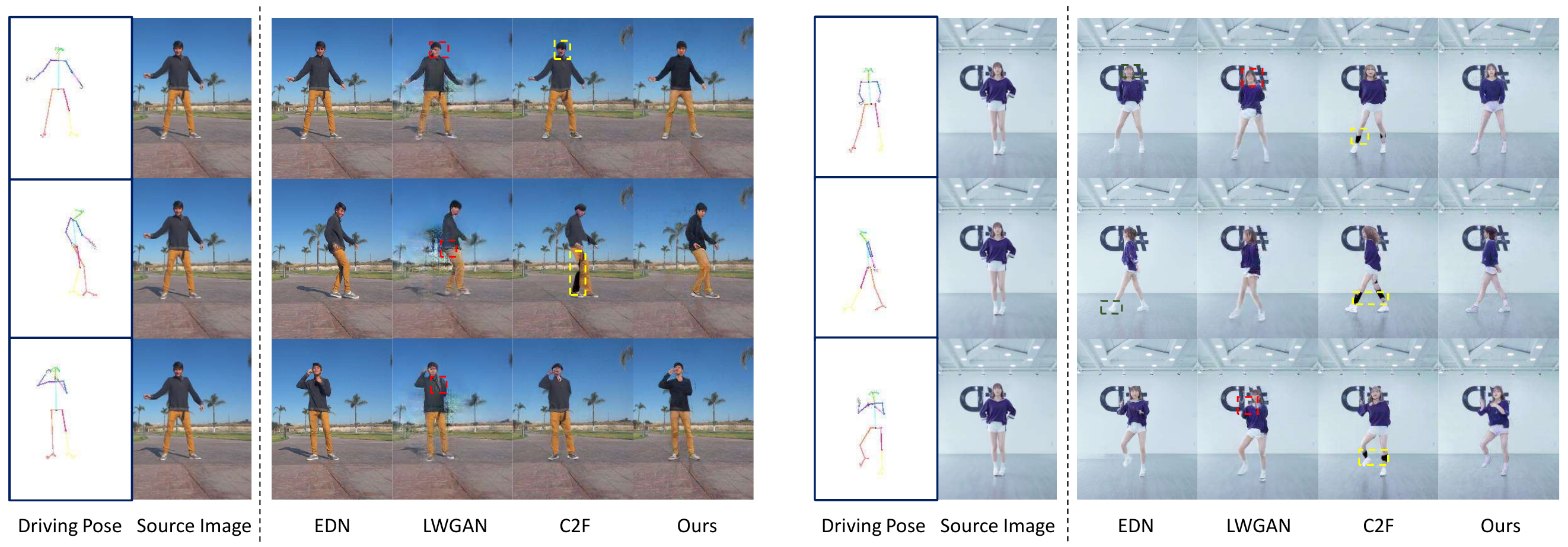}
  \caption{Qualitative results on the SoloDance dataset. 
  %In order to comprehensively compare the generated results, we select three generated images with different poses in each video. 
  Visible artifacts are marked with colored dotted boxes.}
  \label{solo} \vspace{-2mm}
\end{figure*}

A quantitative comparison of our method with other methods on the iPER dataset is shown in Table~\ref{iper_1}. It shows that our proposed method outperforms other methods in terms of image quality and temporal consistency. 
EDN trains the exclusive model for each specific person, so it achieves satisfactory results on all metrics.
Because FSV2V directly generates the whole person image, resulting in unreasonable person images, it performs the worst among all methods.
For LWGAN, there is no constraint on temporal consistency, resulting in jitters in the generated human video. Therefore, LWAGN does not perform well on the TCM metric.
Since half of the videos of rotating actions in the iPER dataset, it is difficult for C2F to handle such a situation with drastic pose differences.
Thanks to proposed region-to-whole strategy, our method can be applied to more situations, thus achieving the best results on all metrics.

Table~\ref{solo_1} provides the comparison of our method with other methods on the SoloDance dataset. It can be seen that we have achieved the best results on the Mask-SSIM, Mask-PSNR, LPIPS and TCM. (Mask means masking the background of the image according to the human parsing map)
This shows that our method has advantages in the quality of generated person images and the temporal consistency of generated videos.
The reason why we don't achieve the best results on other metrics may be that the scale of the SoloDance dataset is not large enough for training the model composed of entirely GANs compared to the iPER dataset. In addition, C2F directly retains the clothes of the source image based on predicted optical flow, which makes it achieve the best results on the FID metric. However, this also leads to generating unreasonable person images in the case of dramatic pose differences between the source person and the driving person. (It can be seen in the areas marked by the yellow dotted boxes of Figures~\ref{iper} and \ref{solo}.) Overall, our method can achieve the state-of-the-art on both two datasets. 

% For Mask-SSIM and Mask-PSNR, it can be seen that the quality of the person images generated by our method has reached the state-of-the-art. 

\vspace{-1mm}
\subsection{Qualitative Results}\label{4.3}
As shown in Figures~\ref{iper} and \ref{solo}, we randomly select three video frames of different poses from synthesized videos on iPER and SoloDance datasets. 
Although EDN can generate realistic images in some cases, unseen poses often lead to unreasonable person images (Marked with green boxes in Figures~\ref{iper} and ~\ref{solo}). 
Therefore, EDN is limited by the pose diversity of the particular person in the training set. 
It can be seen that due to the inaccuracy of the SMPL predicted by HMR~\cite{hmr}, LWGAN will generate uneven texture and jitter (Marked with red boxes in Figures~\ref{iper} and ~\ref{solo}). 
For C2F, when the poses of the driving image and the source image are different dramatically, optical flow prediction becomes inaccurate. 
So the synthesized clothes cannot be warped precisely, which results in unrealistic person images. (Marked with yellow boxes in Figures~\ref{iper} and ~\ref{solo})
This may seriously affect human visual perception.
Compared with other methods, our method can better handle the situation of drastic pose changes while preserving the details. 
In addition, the person images generated by our model generally have clearer faces. 
This is due to our progressive model in which the initial face image provides an important template for the final clear face.

\subsection{Ablation Study}\label{4.4}
In order to verify the roles of the Global Alignment Module (GAM) and Texture Alignment Module (TAM) on the generated results and the effectiveness of the Whole Composition Network (WCN), we perform the ablation experiments on the iPER and SoloDance datasets.
The variants of the framework are as follows:
\begin{itemize}
	\item[$\bullet$] \textbf{RGN w/o GAM.} It means that the foreground image is generated without using the GAM.
    \item[$\bullet$] \textbf{REMOT w/o WCN.} It means that the foreground image is generated through the RGN using the GAM.
    \item[$\bullet$] \textbf{REMOT w/o TAM.} It means that the REMOT model directly adds the feature $F_1^w$ to the feature $F_2^w$ point-to-point, instead of using the TAM.
	\item[$\bullet$] \textbf{REMOT.} It refers to our proposed complete REMOT model.
\end{itemize}

\begin{figure}[t]
  \centering
  \includegraphics[width=\linewidth]{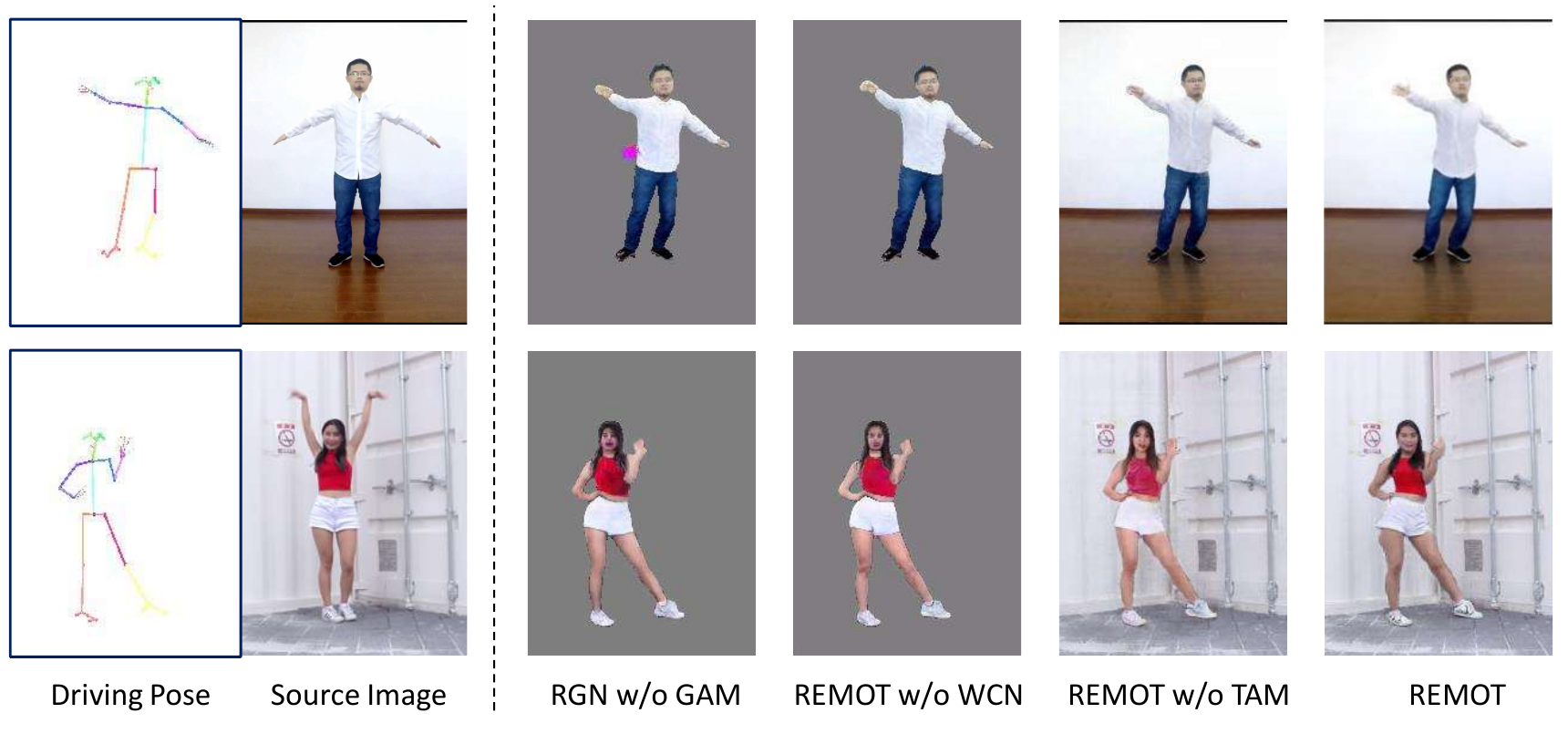}
  \caption{Results of ablation experiments.}
  \label{xiaorong}
  %\vspace{-3mm}
\end{figure}

\begin{table}[t]
  \small
  \caption{Results of ablation experiments of our method on the iPER dataset. * indicates Mask-SSIM and Mask-PSNR.}
  \label{iper_2}
  \begin{tabular}{l|ccccc}
    \toprule
    &SSIM* $\uparrow$&PSNR* $\uparrow$&LPIPS $\downarrow$&FID $\downarrow$&TCM $\uparrow$\\
    \midrule
    RGN w/o GAM     & 0.892 & 20.94  & - & -  & - \\ 
    REMOT w/o WCN   & 0.895 & 21.32  & - & - & - \\
    REMOT w/o TAM   & 0.928 & 26.62  & 0.074 & 55.23  & 0.715 \\ 
    REMOT     & \textbf{0.952} & \textbf{28.87}  & \textbf{0.065}  & \textbf{53.04} & \textbf{0.793} \\ 
    \bottomrule
\end{tabular}
\end{table}

\begin{table}[t]
  \small
  \caption{Results of ablation experiments of our method on the SoloDance dataset.}
  \label{solo_2}
  \begin{tabular}{l|ccccc}
    \toprule
    &SSIM* $\uparrow$&PSNR* $\uparrow$&LPIPS $\downarrow$&FID $\downarrow$&TCM $\uparrow$\\
    \midrule
    RGN w/o GAM    & 0.921 & 24.56  & - & -  & - \\ 
    REMOT w/o WCN  & 0.937 & 25.54  & - & - & - \\    
    REMOT w/o TAM  & 0.941 & 26.25 & 0.053 & 64.37  & 0.723 \\ 
    REMOT  & \textbf{0.953} & \textbf{27.89} & \textbf{0.045} & \textbf{53.29}  & \textbf{0.788} \\ 
    \bottomrule 
\end{tabular}
\end{table}

The experimental results are shown in Tables~\ref{iper_2} and~\ref{solo_2}.
Since the RGN only generates person foreground images, we only calculate the Mask-SSIM and Mask-PSNR. 
Experimental results in Table~\ref{iper_2} show that the GAM does not significantly improve the quality of the generated images on the iPER dataset. 
This is because in iPER videos, the motion of actors is relatively simple and the scale and position of actors remain unchanged, which means that the source image and the driving image are originally in a weak alignment. 
However, the GAM plays a greater role because of the large range of actions of the subjects in the SoloDance dataset. 
Obviously, on both datasets, WCN can significantly improve the quality of generated images. 
Compared with direct feature addition for fusion, TAM can better fuse the features of the generated raw foreground image with the features of the source image. 

As can be seen from Figure~\ref{xiaorong}, WCN plays an important role in enhancing facial details and clothing textures. 
Compared with the direct addition of features, TAM is more beneficial to generate realistic person images. 
%More experimental results can be found in our supplementary material.

\section{Conclusion}
% In this paper, we propose a progressive human motion transfer framework, namely REMOT, which gradually generates person images conditioned on driving poses from region to whole.
% The entire framework abandons the warping operation commonly used in previous works. 
% This strategy can handle the significant noises of the warping flows between images when the source pose and driving pose are different dramatically.
% The region generation strategy also can avoid the difficulty of directly generating the whole person image.
% To generate more realistic images, we propose a Global Alignment Module to transform the foreground of the source image to the proper position and scale for matching the driving person.
% Furthermore, we propose the Texture Alignment Module, which further fuses the features of the source image to refine the raw generated image by computing the similarity between the source image features and the raw generated image features. 
% Experiments on the iPER and SoloDance datasets show that our proposed approach achieves state-of-the-art results.
% % We also conduct extensive ablation studies to demonstrate the effectiveness of our proposed modules.
% In future work, we will try 3D information of the human body for HVMT, which can provide powerful prior knowledge such as 3D shapes and poses of the human body.
In this paper, we propose a progressive human motion transfer framework, namely REMOT, which gradually generates person images from region to whole. 
REMOT abandons the warping operation, which avoids inaccurate flow estimation due to drastic variations of poses.
% The region-to-whole strategy also can avoid the difficulty of directly generating the whole person image.
Compared with directly generating the whole person image, the region-to-whole strategy makes the HVMT task easier and generated person images more realistic.
Moreover, we propose a Global Alignment Module to match the size and position of the source person with those of the driving person.
Furthermore, we propose the Texture Alignment Module to align the features of the source image with the generated image to preserve more details.
Experiments on the iPER and SoloDance datasets show that our proposed approach achieves state-of-the-art results.
In future work, we will try 3D information of the human body for HVMT, which can provide powerful prior knowledge such as 3D shapes and poses of the human body.

\vspace{3mm}
\begin{acks}
This work is supported by the National Key R\&D Program of China under Grand No. 2020AAA0103800, the National Nature Science Foundation of China (62121002, 62022076, U1936210, 62102127), the Fundamental Research Funds for the Central Universities under Grant WK3480000011, the Youth Innovation Promotion Association Chinese Academy of Sciences (Y2021122), and the Hefei Postdoctoral Research Activities Foundation (BSH202101).

\end{acks}

\newpage
\balance
\bibliographystyle{ACM-Reference-Format}
\bibliography{sample-base}
\appendix

\end{document}